\RequirePackage{fix-cm}
\documentclass[smallextended]{svjour3}       
\smartqed  
\usepackage{graphicx}
\usepackage{xcolor}
\usepackage{breakcites}
\usepackage{subcaption}
\usepackage{comment}
\usepackage{amsfonts}

\begin{document}

\title{Multimodal and self-supervised representation learning for automatic gesture recognition in surgical robotics 
}


\author{Aniruddha Tamhane         \and
        Jie Ying Wu \and 
        Mathias Unberath
}


\institute{Aniruddha Tamhane \at
              Johns Hopkins University \\
              \email{atamhan3@jhu.edu}    }       


\institute{All authors are with The Johns Hopkins University, Baltimore, MD.}

\maketitle

\begin{abstract}
Self-supervised, multi-modal learning has been successful in holistic representation of complex scenarios. This can be useful to consolidate information from multiple modalities which have multiple, versatile uses. Its application in surgical robotics can lead to simultaneously developing a generalized machine understanding of the surgical process and reduce the dependency on quality, expert annotations which are generally difficult to obtain. We develop a self-supervised, multi-modal representation learning paradigm that learns representations for surgical gestures from video and kinematics. We use an encoder-decoder network configuration that encodes representations from surgical videos and decodes them to yield kinematics. We quantitatively demonstrate the efficacy of our learnt representations for gesture recognition (with accuracy between 69.6 \% and 77.8 \%), transfer learning across multiple tasks (with accuracy between 44.6 \% and 64.8 \%) and surgeon skill classification (with accuracy between 76.8 \% and 81.2 \%). Further, we qualitatively demonstrate that our self-supervised representations cluster in semantically meaningful properties (surgeon skill and gestures). 
\end{abstract}

\section{Introduction}
\label{intro}

The use of robotic surgery has increased in the recent years. Robotic surgical consoles such as the da Vinci\textsuperscript{\textregistered} Surgical System~\cite{guthart2000intuitive} developed by Intuitive Surgical Inc. (Sunnyvale, CA). are being used in place of traditional surgical tools such as laparoscopes, with the aim of improving surgical quality and patient prognosis. These consoles change the fundamental nature of the surgeon's interaction with the patient. Studying human surgical performance under such a setting is thus a very exciting field of research that is crucial for improving our understanding of surgeon-robot interaction as well as improving patient outcomes. The rich and accurate surgical data available in multiple modes such as videos and kinematics can be used to gain a holistic comprehension of the surgery as well as study human surgical performance closely and accurately.
\\ A lot of recent work in this domain has focused on task-specific, supervised learning from a single modality. We argue that this approach has several drawbacks. For one, a task-specific learning paradigm (such as supervised skill detection, for example) enables a very narrow learning of the overall surgical process, giving very little understanding beyond the specific training task. Further, supervised learning enforces a dependence on expert annotations that are expensive to obtain, subjective and possibly erroneous. Finally, ignoring multiple modalities of information (such as video and kinematics) can be detrimental to learning generalizeable, feature-rich representations. 
\\ We argue that robotic surgery can be broken down into a sequence of gestures or surgemes \cite{varadarajan2009data}. Thus, it is possible to model a surgery similar to language models \cite{mikolov2013distributed, devlin2018bert} over the vocabulary of the surgemes. In this work, we propose a self-supervised learning algorithm that learns task-agnostic surgical gesture 
representations from two modalities (viz. video and kinematics) that can generalize well across multiple tasks.
\\ Our contributions in this paper are as follows: 1) we provide a deep-learning based architecture to learn self-supervised surgical gesture representations from video and kinematics, 2) we quantitatively demonstrate the utility of our representations across multiple tasks by achieving state-of-the-art accuracy in gesture recognition and high accuracy in skill recognition, 3) we qualitatively demonstrate the rich information stored in our representations by visualizing our representations forming semantically meaningful clusters.

\section{Related Work}
We review research literature closely related to our research. This section is organised as follows: we review state-of-the-art supervised learning approaches in surgical robotics in Section \ref{sec:supervised_lit}, unsupervised learning approaches in Section \ref{sec:unsup_lit}, state-of-the-art approaches in video activity recognition in Section \ref{sec:video_lit} and landmark works in self-supervised learning from multimodal sources in \ref{sec:multimodal_lit}. 

\subsection{Supervised Surgical Gesture Recognition}
\label{sec:supervised_lit}
A vast majority of research on surgical robotics learning task focuses on supervised learning. The focus of such work is to learn a specific surgical task such as future prediction, gesture recognition etc from annotated data. In \cite{sarikaya2019surgical}, Sarikaya and Jannin parse the optical flow corresponding to the surgical videos using a Convolutional Neural Network based backbone to perform a supervised surgical gesture classification. They highlight an important point that solely optical flow is a sufficient source of information for learning to classify surgical gestures with a high accuracy. We use this insight to extract the optical flow as a source of domain-independent visual information. In \cite{mazomenos2018gesture}, Mazomenos et al use Recurrent Neural Networks (RNNs) to classify surgical gestures in a supervised manner. Similarly, in \cite{dipietro2016recognizing}, DiPietro et al also applied unidirectional and bi-directional LSTMs for gesture recognition. RNNs, though a natural and effective choice in parsing sequential data such as video frames are computationally expensive and harder to train. In our work, we use CNNs to parse the sequential frames, with each CNN channel dedicated to a particular frame. In \cite{sarikaya2018joint}, Sarikaya et al combine both modalities i.e. optical flow and video for supervised surgeme recognition. This work is also a prime example of integrating multimodal information to improve supervised gesture recognition results.

\subsection{Unsupervised Surgical Gesture Recognition}
\label{sec:unsup_lit}
Unsupervised surgical gesture recognition is a research paradigm with a large potential impact as it does not rely on large amounts of quality, expert annotations like supervised learning.
In \cite{dipietro2018unsupervised}, diPietro and Hager demonstrated that predicting the next surgical gesture using RNNs in an unsupervised manner captures the latent information necessary for surgical gesture recognition. This was demonstrated on the kinematics data in the JIGSAWS dataset. Further, they showed that these embeddings naturally clustered corresponding to distinct higher level activities. In this work, diPietro and Hager have demonstrated the possibility of learning unsupervised representations from surgical data that can perform on par with state-of-the-art supervised learning paradigms on surgical gesture recognition tasks. Further, they demonstrated the versatility of these learnt representations by demonstrating their utility in other tasks such as information retrieval. We however envisage that a unimodal source of information (i.e. kinematics) could potentially have implied a very narrow, task specific learning of representations. In our work, we use a 2D-CNN based encoder-decoder structure to learn representations that encode information from multiple modalities (viz video and kinematics) and perform well across multiple tasks such as gesture recognition, surgeon skill classification and cross-task gesture recognition. The experiments have been explained in detail in Section \ref{sec:expt_res}.

\subsection{Video activity recognition}
\label{sec:video_lit}
Surgical gesture recognition from video data can benefit immensely by borrowing from milestone techniques from video activity recognition. Thus, reviewing key papers in this domain is fruitful to our efforts. In \cite{karpathy2014large}, Karpathy et al review several video classification models (slow, early and late fusion) to provide empirical evidence for the superiority of the slow fusion paradigm. Several works such as \cite{ji20123d} treat videos as 3D videos, and thus extend the notion of a 2D CNN to 3D. This gives us the necessary insights for parsing sequential visual data using CNNs.
Simonyan and Zisserman take a different approach in \cite{simonyan2014two}. They pair each video frame with the corresponding optical flow and parse both through parallel 2D CNN based backbones. This approach has biological justifications, since it has been indicated in \cite{simonyan2014two} that human vision processing occurs in a similar manner, with separate neural pathways dedicated to processing static image and motion information. We have experimented with a two-stream approach using parallel CNN-based video and optical flow parsing streams. However, we finalized upon an encoder-decoder based architecture since it was more suitable for our particular learning task. The architecture is further described in Section \ref{sec: architecture}.
In \cite{girdhar2019video}, Giridhar et al use the transformers-based attention as described in \cite{vaswani2017attention} to learn high attention regions in video frames prior to action recognition. It is worth exploring visual attention based models for surgical activity recognition. We did not explore this further since visual attention was an unnecessary feature for the data provided in the JIGSAWS dataset, since only the surgical instruments (which are the only objects of interest) move in the given video. Thus, we obtain all the information for motion localization from the optical flow.

\subsection{Multimodal self-supervised learning}
\label{sec:multimodal_lit}
Multimodal, self-supervised learning is has shown great promise in learning . In \cite{arandjelovic2017look}, Arandjelovic and Zisserman learnt embeddings for video and audio inputs such that corresponding embeddings for audio and video cluster close to each other. This was achieved by training a two-stream, CNN based network to binary classify embeddings for correspondence. Further, in \cite{arandjelovic2018objects}, they demonstrate that the self-supservised embeddings store information on the source of sound in the video. This information can be extracted by computing a simple correlation between the audio embeddings and different regions of the video frame representation.

\section{Problem Formulation}
\label{sec:1}
Our problem statement is to learn multi-modal representations for surgical robotic activity from video and kinematics data using a self-supervised learning. Formally, given a dataset $\mathbb{D} = (\mathcal{V}_i, \mathcal{K}_i)_{i = 1}^{i = n}$ which is a tuple of surgical videos $\mathcal{V}$ and corresponding kinematics $\mathcal{K}$, we seek a corresponding lower-dimensional representation $r(T(\mathcal{V}_i), \mathcal{K}_i)$, where $T$ is a transformation on $\mathcal{V}$. In our case, $T$ is a function that extracts the optical flow. Further, we wish $r(T(\mathcal{V}_i), \mathcal{K}_i)_{i = 1}^{i = n}$ retain all the critical information pertaining to the surgery, such as the exact surgical gesture, identity of the surgeon and the skill with which the segment of surgery was performed. Self-supervised learning is a natural solution to this problem, given its efficacy in learning holistic information. To this end, we define our problem a maximization of critical information retention in $r$. Thus, our loss criterion $\mathcal{L}$ is a function $\mathcal{L}: \mathbb{R}^d \times \mathbb{R}^d \rightarrow \mathbb{R}$, where $\mathcal{L}$ is some measure of information loss between the information sources i.e. $(\mathcal{V}, \mathcal{K})$ and $r$. Finally, we define our objective as the following empirical loss minimization over the parameters $\theta$:
\begin{equation}
\label{eqn: basic_loss}
     \min_{\theta}\frac{1}{n}\sum_{i = 1}^n \mathcal{L}(r(T(\mathcal{V}_i), \mathcal{K}_i;\theta), \mathcal{V}_i, \mathcal{K}_i)
\end{equation}

Learning the representation function $r$ that encodes information from the surgical videos and kinematics can be achieved using an appropriate proxy learning objective. 
On the basis of our experiments, we propose a 2-D Convolutional Neural Network based encoder-decoder learning objective. 
Further, to learn representations that are independent of video modalities such as luminence, color scheme, contrast etc. we extract the optical flow $\mathcal{O}_i$ from the video frames $\mathcal{V}_i$ through transformation $T$. Finally, we demonstrate that this architecture is helpful in learning representations that outperform the current state-of-the-art in self-supervised surgical gesture recognition. Thus, our problem further reduces to the following: encode information from optical flow in representations $r(T(\mathcal{V}))$ and minimize the information loss $\mathcal{L}$ between the decoded representations $\mathcal{D}(r(T(\mathcal{V})))$ and kinematics $\mathcal{K}$. We empirically choose the loss function $\mathcal{L}$ as the L2 norm of the difference, denoted by $||\cdot||_2$.
\\ Thus, we rewrite the following formulation equivalent to the one described in in Equation \ref{eqn: basic_loss} under the above mentioned assumptions as:
\begin{equation}
\label{eqn: final_loss}
         \min_{\theta, \phi}\frac{1}{n}\sum_{i = 1}^n ||\mathcal{D}({r(T(\mathcal{V}_i); \theta);\phi)} - \mathcal{K}_i||_2^2
\end{equation} 
given an parameterized encoder function $r$ with parameters $\mathcal{\theta}$ and decoder function $\mathcal{D}$ with parameters $\phi$.

\section{Training and Architecture}
\label{sec: architecture}
\begin{figure*}
  \includegraphics[width=\textwidth , height=140pt]{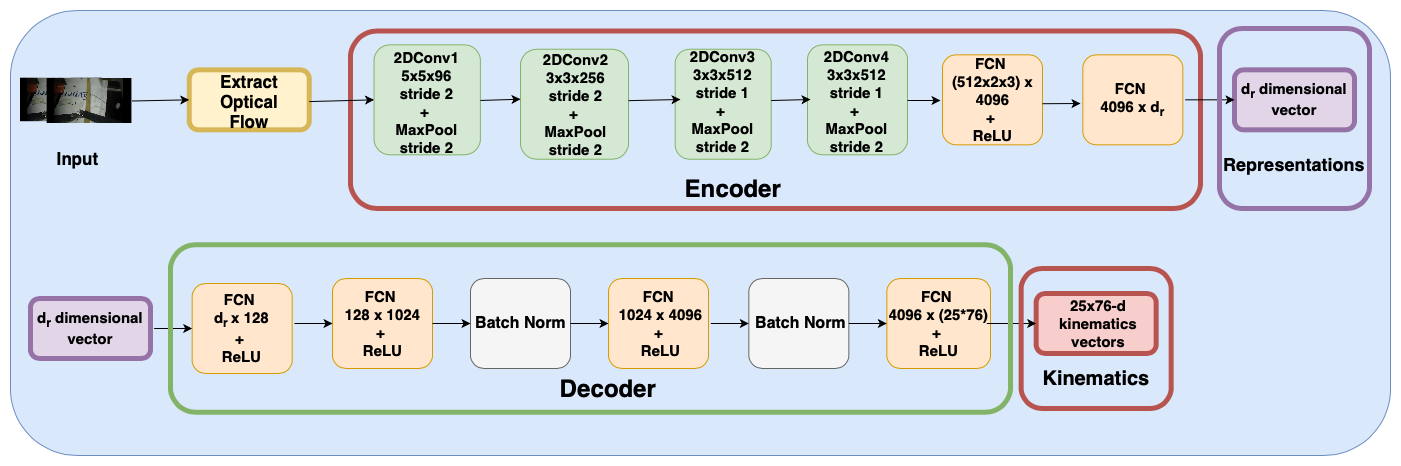}
\caption{Multimodal encoder-decoder network architecture}
\label{fig:1}       
\end{figure*}

We observe that the choice of architecture is key to learning quality representations. We initiated our inquiry with an architecture similar to AVE-Net described in \cite{arandjelovic2018objects}, replacing the optical flow parsing stream with the one we describe in Section \ref{sec:encoder} and experimenting with an RNN-based parser (LSTMs/GRUs) and a 1D CNN based parser for the kinematics stream. We kept the training task identical to the original representations alignment task. We observed that the 1D CNN worked better than the RNN for parsing the kinematics. However, the results were not entirely satisfactory. We observe that similarity in multiple surgical gestures makes the task of identifying a one-to-one mapping from video to kinematics a hard task, thus making the training objective unsuitable for our task. We transformed the training objective from a alignment-based task to an encoder-decoder task, where the objective is to extract the corresponding kinematics vectors from the optical flows. Accordingly, we modified the kinematics decoder to the one described in Section \ref{sec:decoder}. We observe that this configuration worked very well. 
Finally, we choose a simple 2D Convolutional Neural Network based architecture for our Encoder and a fully connected layer based architecture for our decoder $\mathcal{D}$. A visual schematic of the architecture has been provided in Figure \ref{fig:1}. A detailed description has been provided in the subsequent sections.

\subsection{Encoder}
\label{sec:encoder}
We choose a 2D Convolutional Neural Network based architecture to model our encoder function, as shown in Figure \ref{fig:1}. It has been argued in \cite{sarikaya2019surgical} that optical flow is crucial and sufficient for classifying gestures. Further, we argue that optical flow can potentially filter out domain-specific information such as the video quality, contrast, details about the surgical instruments etc. that is not key in understanding the surgical process. Therefore, we add an optical flow extraction as a pre-processing step prior to encoding the representations. We use the Farneback algorithm \cite{farneback2003two} to extract the optical flow and a 2D CNN based encoder backbone similar to the one used by \cite{simonyan2014two} to encode it. Finally, we provide the network a context of roughly 1.67 seconds by sub-sampling the optical flow fields from every alternate frame over 50 frames. Our objective here is to capture a context over a sufficiently large time-frame. 

\subsection{Decoder}
\label{sec:decoder}
We choose a simple FCN with ReLU activations as decoder as shown in Figure \ref{fig:1}. The decoder outputs the 25 kinematics vectors each corresponding to the sampled optical flows. We keep the decoder network relatively shallow to ensure maximum information retention in the representations yielded by the encoder network.

\subsection{Training}
\label{sec:training}
We train our network for an encoder-decoder task wherein the encoder network encodes information from the videos (optical flow) into the representation, which is thereby parsed by the decoder network to extract the corresponding kinematics. This has been elucidated in Section \ref{sec:1}. While training, we uniformly select every alternate frame, thus selecting 25 frames out of 50, yielding a context of ~1.6 seconds given a frame frequency of 30 Hz.  We train our network to encode this set of 25 video frames into representations. We use a Mean Squared Error (MSE) between the kinematics vectors and the decoder output as training loss, as given in Equation \ref{eqn: final_loss}. Using this training methodology, we train our network for 1000 epochs on the Knot Tying, Needle Passing and Suturing datasets available in the JIGSAWS dataset. We observe that the training loss uniformly decreases to with each training epoch, indicating a smooth learning process.

\section{Dataset}
\label{sec: dataset_desc}
We test our learning approach on the JIGSAWS dataset, which contains a series of annotated clips of surgical activity further sub-divided into three datasets (viz knot-tying, needle-passing and suturing) performed using the da Vinci\textsuperscript{\textregistered} Surgical System console \cite{guthart2000intuitive}. The frame-level annotations include surgical gestures, anonymized user (surgeon) identification and surgical skill metrics based on years of surgical experience. The kinematics available on a frame-level consists of a 76-dimensional vector that includes the $x,y,z$ coordinates of the left and right tool tips, the corresponding linear and angular velocities, the rotation matrix and the gripper angle velocities. A detailed description of the JIGSAWS dataset is given in \cite{gao2014jhu}. 

\section{Experiments and results}
\label{sec:expt_res}
We test the quality of our learned representations for a number of tasks as a measure of the holistic understanding of the surgical process. We first divide the frames in each video clip according to the corresponding gestures. We then uniformly sample 25 frames from each set of 50 gesture-frames. Given the sampling frequency of 30 Hz, we encode a gesture context of about a 1.67 seconds. To extract domain-independent information (as explained in Section \ref{sec:encoder}, we extract the optical flow from the sampled video frames. We further sample the 25 corresponding kinematics vectors. 
\\ We train the end-to-end deep encoder-decoder network described in Section \ref{sec: architecture} to encode the representations from the optical flow and decode the kinematics vectors from the representations. To obtain a visual understanding of the information encoded in these representations, we reduce their dimensionality to a 2-dimensional plane using the U-MAP algorithm \cite{mcinnes2018umap}. We choose U-MAP in particular because of its computational efficiency and ability to preserve global structure, which would be crucial to appreciating the differences in surgical gestures. We then perform a gesture classification experiment using a  gradient boosting based classifier trained solely on the representations. Further, we also perform a skill-classification experiment using another gradient-boosting based classifier. Finally, we perform a gesture classification experiment on a set of cross-task representations i.e. representations generated for a particular surgical task (eg: suturing) using an encoder trained on a different surgical task (eg: knot-tying). Further details and results of the experiments have been discussed in the following sub-sections.

\subsection{Representations visualizations using U-MAP}
\label{sec:u_map_rep}
\begin{figure}
\centering
\includegraphics[scale=0.5]{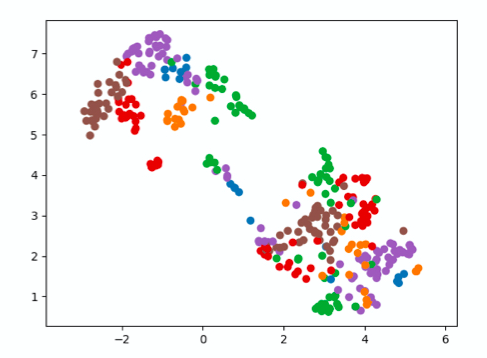}
\captionof{figure}{Knot Tying representations (gesture)}
\label{fig:kt_gesture}
\end{figure}

\begin{figure}
\centering
\includegraphics[scale=0.5]{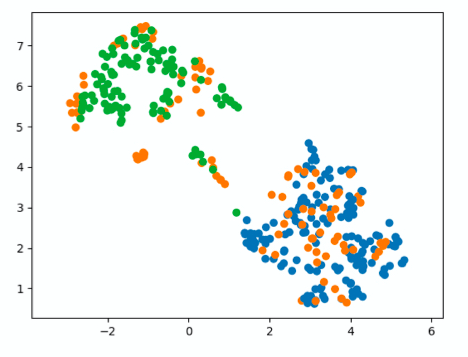}
\captionof{figure}{Knot Tying representations (skill)}
\label{fig:kt_skill}
\end{figure}

\begin{figure}
\centering
\includegraphics[scale=0.5]{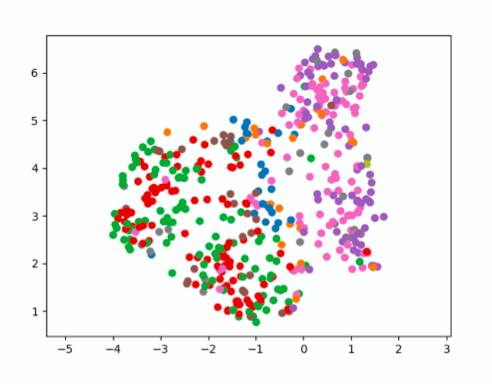}
\captionof{figure}{Needle Passing representations (gesture)}
\label{fig:np_gesture}
\end{figure}

\begin{figure}
\centering
\includegraphics[scale=0.5]{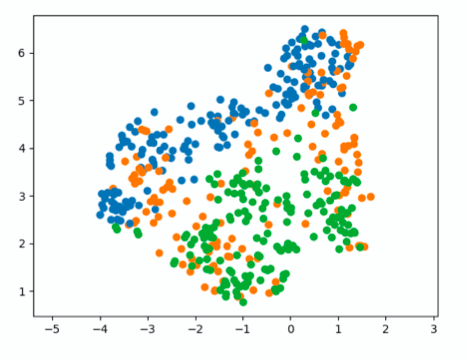}
\captionof{figure}{Needle Passing representations (skill)}
\label{fig:np_skill}
\end{figure}

\begin{figure}
\centering
\includegraphics[scale=0.5]{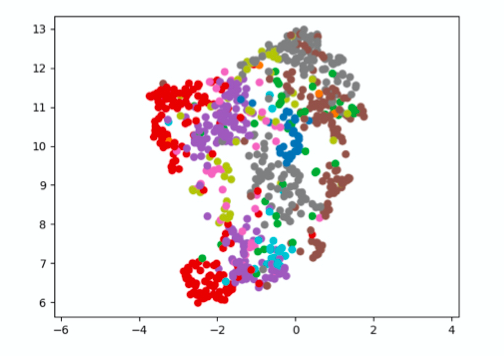}
\captionof{figure}{Suturing representations (gesture)}
\label{fig:sut_gesture}
\end{figure}

\begin{figure}
\centering
\includegraphics[scale=0.5]{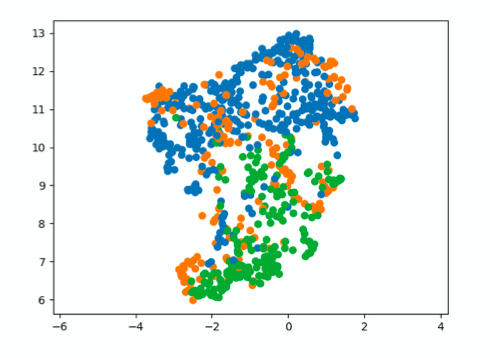}
\captionof{figure}{Suturing representations (skill)}
\label{fig:sut_skill}
\end{figure}

We use the U-MAP algorithm to perform a dimensionality reduction on the learnt representations. We choose U-MAP in particular because of its property to preserver local as well as global distances in the lower dimensional latent-space. We plot the two-dimensional projections of the representation from the Knot Tying dataset, color-coded for different gestures in Figure \ref{fig:kt_gesture} and for different skill-levels in Figure \ref{fig:kt_skill}. We generate similar plots for the Needle Passing dataset in Figures \ref{fig:np_gesture}, \ref{fig:kt_skill} and Suturing dataset in Figures \ref{fig:sut_gesture}, \ref{fig:sut_skill}. It is interesting to observe that the in each of the skill-based plots (Figures \ref{fig:sut_skill}, \ref{fig:np_skill}, \ref{fig:kt_skill}), the representations neatly cluster into two distinct skill-based clusters corresponding to "beginner" and "expert" skill level, with the representations from the "intermediate" skill level spread across both these clusters. A possible explanation for this phenomenon is that the skill "intermediate" is a vague category and encompasses people with varying skill-levels that may be closer to the "beginners" or the "experts". Another interesting observation from the gesture-based plots is that the gestures-based plots (Figures \ref{fig:sut_gesture}, \ref{fig:np_gesture}, \ref{fig:kt_gesture}) is that each individual gesture (denoted by a different color) has two distinct clusters corresponding to a distinct skill category. Thus, it is evident from this visualization that each gesture has a unique representation depending on whether it has been performed by an expert or a beginner.

\subsection{Skill recognition}
\label{sec:skill_rec}
We test the efficacy of our representations in learning surgeon skill levels. In particular, we train the XGBoost, a gradient boosting based classifier introduced in \cite{chen2016xgboost} on the representations for a 3-class classification task on the self-reported user skill.  We do not train the encoder while training the classifier. We repeat our experiment for 5 randomly chosen train-test splits. We report our findings as an aggregate of the five random splits in Table \ref{tab: skill_results} (in the format of mean $\pm$ standard deviation). We observe that the results indicate a significant retention of information related to surgeon skill. Further, a major portion of the classification error is contributed by surgeons having an "intermediate" skill. This reaffirms the observation made in Section \ref{sec:u_map_rep} that the embeddings falling in this particular skill category follows a uniform spread from "beginner" to "expert" and does not necessarily cluster tightly, unlike the other two skill levels. This makes it significantly harder to classify.
\\ \\
\begin{table}[]
\caption{Skill recognition results}
\label{tab: skill_results}
\begin{tabular}{lllll}
\hline\noalign{\smallskip}
Dataset & Accuracy & Precision & Recall & F-1   \\
\noalign{\smallskip}\hline\noalign{\smallskip}
Knot-Tying     & 0.768 $\pm$ 0.0303    & 0.766 $\pm$ 0.0336    & 0.768 $\pm$ 0.0303  & 0.758 $\pm$ 0.0311 \\
Needle-Passing & 0.808 $\pm$ 0.0335   & 0.808 $\pm$ 0.0335     & 0.808 $\pm$ 0.0335  & 0.800 $\pm$ 0.03535 \\
Suturing       & 0.812 $\pm$ 0.0228    & 0.810 $\pm$ 0.0254    & 0.812 $\pm$ 0.0228  & 0.804 $\pm$ 0.0267\\
\noalign{\smallskip}\hline
\end{tabular}
\end{table}

\subsection{Gesture recognition}
\label{sec:gest_rec}
We test the efficacy of our representations in learning different gestures from the videos. Similar to the skill recognition task described in Section \ref{sec:skill_rec}, we train an XGBoost classifier on the representations for gesture classification. We freeze the encoder weights while training the classifier. We repeat our experiment for 5 randomly chosen train-test splits. We report our findings as an aggregate of the five random splits in Table \ref{tab: gesture_results}. We observe that the results are promising indicating a good retention of gesture related information in the representations. Further, we observe that quantitatively, our gesture classification results outperform those reported in \cite{dipietro2018unsupervised}. We attribute this to the fact that our representations encode multimodal information from both kinematics and video streams.

\begin{table}[]
\caption{Gesture recognition results}
\label{tab: gesture_results}
\begin{tabular}{lllll}
\hline\noalign{\smallskip}
Dataset        & Accuracy & Precision & Recall & F-1   \\
\noalign{\smallskip}\hline\noalign{\smallskip}
Knot-Tying     & 0.762 $\pm$ 0.0487    & 0.774 $\pm$ 0.0531    & 0.762 $\pm$ 0.0487 & 0.758 $\pm$ 0.0496 \\
Needle-Passing & 0.696 $\pm$ 0.0328   & 0.694 $\pm$ 0.0376    & 0.696 $\pm$ 0.0305  & 0.674 $\pm$ 0.0351 \\
Suturing       & 0.778 $\pm$ 0.0239   & 0.782 $\pm$ 0.0259    & 0.778 $\pm$ 0.0239  & 0.754 $\pm$ 0.0261 \\
\noalign{\smallskip}\hline
\end{tabular}
\end{table}

\subsection{Transfer learning for gesture recognition}
We measure the robustness of our learned algorithm through a transfer-learning based gesture recognition task. We generate representations using an encoder network trained on one dataset (e.\,g.~: Knot Tying) and use them for gesture classification on the other two datasets (Needle Passing and Suturing in this case) similar to the gesture recognition task described in Section \ref{sec:gest_rec}. We repeat the gesture classification experiment for 5 randomly chosen train-test splits. As in the original gesture classification experiment, we train an XGBoost classifier on the representations generated using the encoder initialized with transferred weights. We do not retrain the encoder while training the classifier. We report our findings as an aggregate of the results of the 5 random split based experiments as follows: results obtained using an encoder trained on the Needle Passing dataset are reported in Table \ref{tab: np_gesture_results}, the Knot Tying dataset are reported in Table \ref{tab: kt_gesture_results} and the Suturing dataset are reported in \ref{tab: sut_gesture_results}. We observe that there is a slight decrease in performance after transfer learning. However, the results are comparable to the baseline results and those obtained from other state-of-the-art unsupervised surgical learning works. This indicates that promising results can be achieved by enabling end-to-end training (including training the encoder) during the task-specific learning.

\begin{table}[]
\caption{Transfer learning (from Needle Passing) for gesture recognition}
\label{tab: np_gesture_results}
\begin{tabular}{lllll}
\hline\noalign{\smallskip}
Dataset    & Accuracy & Precision & Recall & F-1   \\
\noalign{\smallskip}\hline\noalign{\smallskip}
Knot-Tying & 0.568 $\pm$ 0.03271   & 0.608 $\pm$ 0.05403 & 0.568 $\pm$ 0.03271  & 0.570 $\pm$ 0.03605 \\
Suturing   & 0.598 $\pm$ 0.01923   & 0.570 $\pm$ 0.02738  & 0.598 $\pm$ 0.01923  & 0.557 $\pm$ 0.02509 \\
\noalign{\smallskip}\hline
\end{tabular}
\end{table}

\begin{table}[]
\caption{Transfer learning (from Knot Tying) for gesture recognition}
\label{tab: kt_gesture_results}
\begin{tabular}{lllll}
\hline\noalign{\smallskip}
Dataset    & Accuracy & Precision & Recall & F-1   \\
\noalign{\smallskip}\hline\noalign{\smallskip}
Needle-Passing & 0.446 $\pm$ 0.05319   & 0.464 $\pm$ 0.06188 & 0.444 $\pm$ 0.05176  & 0.430 $\pm$ 0.05148 \\
Suturing   & 0.624 $\pm$ 0.02302   & 0.634 $\pm$ 0.04560  & 0.624 $\pm$ 0.02302  & 0.590 $\pm$ 0.02345 \\
\noalign{\smallskip}\hline
\end{tabular}
\end{table}

\begin{table}[]
\caption{Transfer learning (from Suturing) for gesture recognition}
\label{tab: sut_gesture_results}
\begin{tabular}{lllll}
\hline\noalign{\smallskip}
Dataset    & Accuracy & Precision & Recall & F-1   \\
\noalign{\smallskip}\hline\noalign{\smallskip}
Knot-Tying & 0.648 $\pm$ 0.04338   & 0.648 $\pm$ 0.04338 & 0.648 $\pm$ 0.04338  & 0.650 $\pm$ 0.04062 \\
Needle-Passing   & 0.448 $\pm$ 0.04658   & 0.458 $\pm$ 0.03898  & 0.446 $\pm$ 0.04505  & 0.434 $\pm$ 0.04605 \\
\noalign{\smallskip}\hline
\end{tabular}
\end{table}

\section{Discussion and future work}
Surgical robotics is an exciting area for applying multi-modal, self-supervised learning techniques due to the complexity of the task, multiple modes of information and lack of dependence on expert annotated data. The choice of the how to achieve the self-supervised learning can be varied. Previous works in literature have used varied proxy learning objectives (that we denote by $\mathcal{L}$ in Equation \ref{eqn: basic_loss}). For example, the sequential nature of the video/kinematics frames has been utilized to devise a future prediction task in \cite{dipietro2018unsupervised}, where the learned representations were used in gesture recognition and information retrieval. In another example, the audio-visual correspondence task as described in \cite{arandjelovic2018objects, arandjelovic2017look} has been used to improve audio classification and enable sound localization in videos. This can be transformed into a kinematics/visual correspondence task to learn the surgical representations. We observe that this choice is sensitive to the dataset in and the nature of information that is expected from the learnt representations. For example, we observe that the video-kinematics alignment task using triplets loss as described in \cite{arandjelovic2018objects} is a bad learning objective for surgical robotics data as provided in the JIGSAWS dataset because surgery involves a repetition of similar gestures that have similar kinematics. This creates a problem where there is no easy one-to-one correspondence between the video and kinematics sets, rendering the triplet loss based training to be very difficult. A possible way to circumvent this problem is to generate negative samples in each triplet that are sufficiently distant from the base sample by some distance metric. Further, we observe that training RNNs takes more computational resource and cannot be easily parallelized as compared to training CNNs. Hence, we focus on using an architecture primarily built using CNNs and fully-connected networks.
\\ We introduce an encoder-decoder based self-supervised learning system that transforms video stream into corresponding kinematics, as explained in detail in Sections \ref{sec: architecture}, \ref{sec:1} . We further demonstrate empirically that the learnt representations effectively encode information on surgical gestures and surgeon skill. Also, the representations are sufficiently robust across multiple tasks, thus facilitating transfer learning. This also raises the possibility that using an end-to-end, task-specific training methodology can further improve the accuracy in all the mentioned tasks. Finally, we visually observe that the representations form clusters corresponding to a specific gesture and skill-level. 
\\ There are a few limitations to our work. One being that JIGSAWS dataset has a single camera perspective in all the videos, thus making it impossible to test the learning of the camera transformation matrix. Secondly, our encoder network does not parse 3D visual information, which has been provided in the JIGSAWS dataset through a left and right camera feed for each video. This could possibly lead to learning information on new tasks such as depth perception and also simultaneously improve accuracy on the currently existing tasks. Finally, while our work does solve the surgeon skill classification task, it does not address the specific differences in surgical behaviour (e.\,g.~ differences in kinematics) that could potentially address the gap in surgical skill and be used as milestones in surgeon training.
\\There also are additional applications of the proposed representation that future work could explore, such as future task prediction, information retrieval, surgical activity segmentation etc. Training these representations on a sufficiently large and diverse surgical dataset can potentially lead to a standardized architecture for parsing surgical activity that is highly transferable across datasets and tasks. This could in turn enable a universal technology that tracks surgical progress in real-time, giving feedback regarding possible mistakes, surgical scene depth, next gesture suggestion etc. with a high accuracy. 


%
%

\bibliographystyle{spmpsci}      

 \bibliography{references}

\end{document}